\documentclass{article}

\usepackage[numbers]{natbib}
\usepackage[preprint]{neurips_2022}

\usepackage[utf8]{inputenc} 
\usepackage[T1]{fontenc}    
\usepackage{hyperref}       
\usepackage{url}            
\usepackage{booktabs}       
\usepackage{amsfonts}       
\usepackage{nicefrac}       
\usepackage{microtype}      
\usepackage{xcolor}         
\usepackage{enumitem}

\title{What is my math transformer doing?  \newline Three results on interpretability and generalization}

\author{%
 Fran\c{c}ois Charton \\
 Meta AI\\
}

\begin{document}

\maketitle

\begin{abstract}
This paper investigates the failure cases and out-of-distribution behavior of transformers trained on matrix inversion and eigenvalue decomposition. I show that incorrect model predictions still retain deep mathematical properties of the solution (e.g. correct eigenvalues, unit norm of eigenvectors), and that almost all model failures can be attributed to, and predicted from, properties of the problem or solution. This demonstrates that, when in doubt, math transformers do not hallucinate absurd solutions (as was sometimes proposed) but remain ``roughly right''. I also show that the careful choice of a training dataset can accelerate training, while allowing the model to generalize out of its training distribution, invalidating the idea that transformers ``merely interpolate'' from memorized examples.
\end{abstract}

\section{Introduction}

Transformer-based AI for mathematics is a fast-developing field. Over recent years, transformers were applied to a wide range of problems: arithmetic \citep{nogueira2021investigating}, linear algebra \citep{LAWT} , polylogarithm identities \citep{dersy2022}, symbolic integration \citep{lample2019deep}, symbolic regression \citep{biggio2021neural}  and theorem proving \citep{polu2020generative}.  Meanwhile, limitations of transformers were found, which may restrict their use in maths and science. In this paper, I challenge three commonly discussed limitations, namely:  

\begin{itemize}[noitemsep,nolistsep]
\item that transformers are black boxes, and there is no way to know {\bf how} they solve a problem. In mathematics, this means one cannot tell whether the model has learned the abstract concepts needed to solve the problem, or just interpolates between memorized training examples.
\item that transformers have no sense of the correctness of their results. They sometimes hallucinate absurd solutions, instead of remaining ``roughly right'' or admitting failure. 
\item that trained transformers are brittle, and struggle with out-of-domain generalization.
In mathematics, the procedure used to generate the training data heavily influences the problems that the model can solve accurately.
\end{itemize}

Experimenting with three problems of linear algebra, eigenvalue calculation, diagonalisation and matrix inversion, in the setting described by \cite{LAWT}, I show that mathematical properties are indeed learned by transformers, and that their failure cases can be understood and predicted. I also show that by carefully selecting the training dataset, I can improve model performance and generalize far away from the training distribution, challenging the idea that transformers ``merely interpolate''.

\section{What is my model doing? Learning the spectral theorem.}\label{sec:explain}
 
In the {\bf diagonalization task} (``eigenvectors'' in \cite{LAWT}), a model is trained to decompose a symmetric $5 \times 5$ matrix $M$, by predicting a vector $\Lambda \in \mathbb{R}^5$ (with $\lambda_1 \geq \lambda_2 \geq \dots \geq \lambda_5$) and a $5\times 5$ matrix $H$ such that $H^T M H = \textrm{diag}(\Lambda)$. Theory~\cite{golub13} tells us that the coordinates of $\Lambda$ are the eigenvalues of $M$, and the columns of $H$ the corresponding eigenvectors. Besides, $H$ is orthogonal, that is, $H^{-1}=H^T$, or, equivalently, all its rows and columns have unit norm and are mutually orthogonal. Because its coordinates are sorted, $\Lambda$ is unique. The columns of $H$, on the other hand, are defined up to a sign change (or a transformation from the symmetry group $O(k)$ when $k$ eigenvalues are equal).

As in \cite{LAWT}, a sequence-to-sequence transformer (see appendix~\ref{app:architecture} for details) is trained to predict the decomposition $(\Lambda, H)$ of a matrix $M$. During training,  the model minimizes the cross-entropy between its predictions and the sequences representing $\Lambda$ and $H$. At test time, model accuracy is defined as the quality of the diagonalisation, i.e. whether $\| H^T M H - \Lambda \| / \| \Lambda \| < \tau$ (using the $L^1$ norm, and with tolerance $\tau=5\%$). In this experiment, the model is trained from examples only, and no problem-specific inductive bias is introduced, either in the architecture or in the training procedure. To determine if some of the theoretical properties of diagonalization are learned, I run the trained model on a test set of 50000 random matrices, and investigate its predictions. 

The model achieves an accuracy of $92.0\%$. However, in $99.9\%$ of the test cases, the eigenvalues of the input matrix $M$ are predicted with less than $1\%$ relative error (in $L^1$ norm), and within $0.5\%$ in $96.1\%$ of test cases. Also, in $98.9\%$ of the test cases, the norms of all rows and columns in the predicted $H$ are in the interval $[0.99, 1.01]$, as theory dictates. These two mathematical properties of diagonalization, i.e. that $\Lambda$ is the eigenvalues, and that the columns of $H$ have unit norm, have been learned by the model. They are verified {\bf even in incorrect predictions}.

In this experiment, the model achieves high in-domain accuracy, but similar results are observed in weaker models. On a ``half-trained'' model that only achieves $70\%$ accuracy, the eigenvalues are predicted  (within $1\%$) in $99.6\%$ of the test cases, and all rows and columns have unit norms in $96.7\%$. For larger matrices ($6\times 6$), the model achieves a meager $43\%$ accuracy. Yet, eigenvalues are predicted within $1\%$ in $99.6\%$ of the test cases, and rows and columns of $H$ have unit norm in $93.1\%$. 

Theory predicts that the rows and columns of $H$ should be orthogonal. This property can be quantified by computing the dot products between sucessive normalized rows and columns of $H$. The dot products are second order approximations of the difference between $\pi/2$ and the angle between vectors (which should be zero if $H$ is orthogonal). On the test set, all angles are within $0.1$ radians ($5.7^\circ$) of $\pi/2$ in $95.2\%$ of test cases, and $0.05$ radians ($2.9^\circ$) in $93.6\%$. The lack of orthogonality between lines and columns accounts for almost all failure cases: in $99.5\%$ of successful model predictions, all angles between successive rows and columns are less than $0.03$ radians, and $H$ is close to orthogonal. On the other hand, one angle is larger than $0.03$ radians in $90\%$ of model failures. 

These experiments teach us three lessons about math transformers. First, deep mathematical properties are learned during training: all eigenvalues are correctly predicted, and all columns of $H$ have unit norms, even when the model fails to predict the correct diagonalisation, and even for models with low accuracy (half-trained, or trained on harder problems). Second, math transformers do not seem to hallucinate absurd solutions. Even when the model fails, $\Lambda$ is correct, and $H$ is close to orthogonal. Finally, they provide a simple mathematical explanation for almost all model failures.

\section{Predicting failure: verifiers for math transformers}\label{sec:failures}

On the diagonalization task, almost all incorrect model predictions can be attributed to $H$ not being orthogonal. From this observation, a useful statistic for predicting model failure can be derived: the condition number of $H$ (i.e. the ratio of its largest and smallest singular values, henceforth $c(H)$). When $H$ is orthogonal, we have $c(H) = 1$ (else $c(H)>1)$. Over the 50 000 test cases, correct model predictions have an average condition number of $1.01$ (with a standard deviation of $0.0065$). For model failures, the average condition number is $1.28$. Using the rule $c(H) < 1.045$, $99.3\%$ of model successes and failures can be predicted. More precisely, we have $c(H)<1.045$ in $99.94\%$ of correct predictions, and $c(H)>1.045$ in $96.7\%$ of model failures. 

A similar situation arises for $5\times 5$ matrix inversion. Over a test set of 50 000 examples, a transformer has an accuracy of $89.0\%$. As in \cite{LAWT}, accuracy is defined by how close the product of the model prediction $P$ and the input matrix $M$ is to identity, i.e. $\| PM - I \| / \| I \| < \tau$ ($\tau=5\%$). But we can also compute the $L^1$ distance between the model prediction and the inverse $\|P-M^{-1}\| / \| M^{-1} \|<\tau$. On this metric, accuracy is $98.2\%$ with $5\%$ tolerance, and $99.6\%$ with $25\%$. When in doubt, the model does not hallucinate, but provides a rough approximation to the correct solution $M^{-1}$.

This provides us with a complete mathematical explanation of model failure for the inversion task. Whereas the model fails on $11\%$ of test cases, its predictions are within $5\%$ of the correct solution in $98.2\%$, and in $84\%$ of failures ($(98.2 - 89)/11$). 
In such cases, the model predicts an approximation of $M^{-1}$ that turns out not to be a ``good inverse'' of $M$. We know from theory that this happens when $M$ has a large condition number $c(M)$, and therefore we can use $c(M)$ to predict model failure. On the test set, the matrices correctly inverted by the model have an average condition number of $15.8$ (with a standard deviation of $13.3$). For model failures, the average condition number is $640.5$. The decision rule $c(M)<62$ predicts model success in $98.0\%$ of cases, and we have $c(M)<62$ in $99.0\%$ of correct predictions, and $c(M)>62$ in $89.8\%$ of failures. Note that for this task, we do not even need to run the model, since success can be predicted from its input $M$ only.

These experiments indicate that verifiers, external routines that can predict a model success from its input or output, can be computed from problem-specific statistics. In linear algebra, this is of little practical interest because model predictions can be checked in a few matrix multiplications. Verifiers, however, are important in some areas of mathematics (e.g. theorem proving).

\section{Out-of-domain generalization and the role of generators}\label{sec:ood}

On the eigenvalue computation task, I have shown, in~\cite{LAWT}, that models trained on Wigner matrices (with eigenvalues distributed as a semicircle law) do not generalize to test sets with different distributions of eigenvalues (uniform, Gaussian, Laplace, or positive). On the other hand, models trained on matrices with Laplace distributed eigenvalues (Laplace models, henceforth) generalize to all test sets. 

Table~\ref{tab:ood55} presents additional results for seven eigenvalue distributions (semi-circle, uniform, Gaussian, Laplace, absolute-semicircle, absolute-Laplace, and Marchenko-Pastur, see Appendix~\ref{app:generation}). In the first four, eigenvalues are symmetrically distributed around zero. In the last three, all eigenvalues are positive. Also, the semicircle, uniform, absolute semicircle and Marchenko-Pastur  distribution have bounded support, whereas the Gaussian, Laplace and absolute Laplace allow for large eigenvalues.

\begin{table}[h]
    \small
    \centering
    \begin{tabular}{lccccccc}
        \toprule
        \small
        & Semi-circle &  Uniform & Gaussian & Laplace & abs-sc & abs-Lapl& Marchenko\\
        \midrule
        Semi-circle & 100 & 34 & 36 & 39 & 1 & 5 & 0 \\
        Uniform & 93 & 100 & 76 & 70 & 92 & 70 & 2 \\
        Gaussian &  100 &  100 &  100 &  100 &  100 &  100 & 99 \\
        Laplace &   100 &  100 &  100 &  100 &  100 &  100 &  100 \\
        \midrule
        Abs-semicircle & 0 & 5 & 4 & 4 & 100 & 78 & 20 \\
        Abs-Laplace&  0 & 4 & 5 & 5 & 100 & 100 & 100 \\
        Marchenko-Pastur & 0 & 4 & 4 & 4 & 100 & 76 & 100\\ 
        
       \bottomrule
    \end{tabular}
    \smallskip
    \caption{\small \textbf{Out-of-distribution generalization. Eigenvalues of 5x5 matrices}. Rows are the training distributions, columns the test distributions.}
    \label{tab:ood55}
\end{table}

The Wigner ensemble, the obvious default choice for random matrices, turns out to be the worst for out-of-distribution generalization. On the other hand, the Gaussian or Laplace models generalize to all test sets. Models trained on positive eigenvalue distributions do not generalize to symmetric (non-positive) test distributions, because negative eigenvalues were never encountered during training (the $4$ to $5\%$ performance achieved by positive models on the Laplace, Gaussian and Uniform ensembles roughly corresponds to the number of positive matrices in the test set). But models trained on symmetric distributions can generalize to positive matrices. Finally, it is interesting to note that models trained on distributions with compact support (semi-circle, uniform, abs-semicircle and Marchenko-Pastur) generalize less well than their unbounded counterparts.

Besides generalizing better, the Laplace and Gaussian models are more data efficient. To achieve $99\%$ accuracy on a Wigner (semi-circle) test set, the Gaussian model needs $2.4$ million training examples, the Laplace model $2.7$ and the semi-circle model $3.6$. On a test set of positive matrices, the Gaussian and Laplace model achieve $99\%$ accuracy in $2.1$ and $2.4$ million examples, the positive model in $3.9$ million (see Table~\ref{tab:oodspeed} in Appendix~\ref{app:eigenvalues}). 
As problem dimension increases, so does the advantage of Gaussian and Laplace models. On $8 \times 8$ matrices (Table~\ref{tab:ood810}), Gaussian and Laplace models achieve $99\%$ accuracy on a semi-circle test set after $11.4$ and $13.2$ million examples. After $36$ million examples, our best uniform and semicircle models only achieve $91$ and $0.5\%$ accuracy. With deeper encoders (8 and 12 layers), the Laplace and Gaussian models can predict the eigenvalues of $10\times 10$ Wigner matrices with $100\%$ accuracy (in $12.9$ and $23.1$million examples, larger models allow for faster learning). The best (semicircle) models reported in \cite{LAWT} only achieve $25\%$ accuracy after $360$ million examples. 

\begin{table}[ht]
    \small
    \centering
    \begin{tabular}{lccccccc}
        \toprule
        \small
        & Semi-circle &  Uniform & Gaussian & Laplace & abs-sc & abs-Lapl& Marchenko\\
        \midrule
        8x8 matrices\\
        Semicircle & 0& 0& 0& 0& 0& 0 &0 \\
        Uniform & 91 & 100 & 65 &57 & 89 & 55 & 0 \\
        Gaussian & 100 & 100 &100  &99  &100  &99  & 41  \\
        Laplace & 100 & 100 & 100 & 100 & 100 & 100 & 97  \\
        Abs-semicircle & 0& 1 & 1 & 0 & 100 & 53 & 0 \\
        Abs-Laplace & 0 & 1& 1  & 1 & 100 & 100 & 98 \\
        Marchenko-Pastur & 0 & 0 & 0 & 0 & 1 & 1 & 20 \\
        \midrule
        10x10 matrices \\
        Gaussian (12/1 layers) & 100 & 100 & 100  & 98 & 100 & 97 & 3 \\
        Laplace (8/1 layers) & 100 & 100 & 100 & 100 & 100 & 100 & 74 \\
 
       \bottomrule
    \end{tabular}
    \smallskip
    \caption{\small \textbf{Out-of-distribution generalization. Eigenvalues of 8x8 and 10x10 matrices, accuracy after 36 million examples.} Rows are the training distributions, columns the test distributions.}
    \label{tab:ood810}
\end{table}

Achieving $100\%$ accuracy on test sets of positive matrices, with Laplace or Gaussian models, rules out the idea that transformers interpolate between memorized examples. For $8\times 8$ and $10\times 10$ matrices, there is almost no overlap between the training and test sets: the probability of a Gaussian or Laplace matrix having only positive eigenvalues is $0.4\%$ and $0.1\%$ respectively.

I obtain similar results when diagonalizing $5\times 5$ matrices (Table~\ref{tab:oodev55}). After training on 80 million examples, the best models achieve $94\%$ accuracy on the semicircle test set. As with the eigenvalue task, the semicircle model does not generalize out of distribution, and the Gaussian and Laplace generalize to all test distributions, and achieve about $80\%$ accuracy. Previous observations on data efficiency also apply: on the semicircle test set, the Laplace and Gaussian models need $37$ and $45$ million examples to achieve $90\%$ accuracy, whereas the semicircle model needs $50$ million (see Table~\ref{tab:oodspeed2} in Appendix~\ref{app:eigenvalues}). 

\begin{table}[h]
    \small
    \centering
    \begin{tabular}{lccccccc}
        \toprule
        \small
        & Semi-circle &  Uniform & Gaussian & Laplace & abs-sc & abs-Lapl & Marchenko\\
        \midrule
        Semicircle & 93& 15& 18 & 18& 0& 0 &0 \\
        Uniform & 91 & 80 & 62 &56 & 81 & 50 & 2 \\
        Gaussian & 94 & 80 & 81  & 77  & 84 & 69 & 80  \\
        Laplace & 94 & 79 & 81 & 78 & 84 & 70 & 81 \\
        Abs-semicircle    & 0 & 3  & 2 & 2 & 82 & 51 & 15 \\
        Abs-Laplace        & 0 & 2 & 3  & 3 & 79 & 71 & 82 \\
        Marchenko-Pastur & 0 & 1 & 2 & 2 & 64 & 42 & 88 \\
 
       \bottomrule
    \end{tabular}
    \smallskip
    \caption{\small \textbf{Out-of-distribution generalization. Diagonalization of 5x5 matrices}. Rows are the training distributions, columns the test distributions.}
    \label{tab:oodev55}
\end{table}
Finally, experiments with symmetric matrix inversion (Appendix~\ref{app:oodinverse}) confirm that Gaussian and Laplace distributions generalize better, and that models trained on positive matrices only generalize to positive test sets. This suggests that the choice of a good training distribution might not be task-specific, and that some distributions may generalize out-of-domain for a large class of problems.

\section{Conclusion}

Experimenting with three problems of linear algebra, I have shown that transformers can learn mathematical properties: all their predictions, correct or not, satisfy some properties (correct eigenvalues and unit vectors for diagonalization). Also, model failures do not happen at random, and can be predicted from the input or the predicted solution. Finally, I show that selecting an appropriate training set improves both out-of-distribution generalization, and model performance and data efficiency. 
These experiments were designed by leveraging the mathematical theory of random matrices and linear algebra. This demonstrates how mathematical problems can be used as frameworks for understanding transformers, trying to explain their predictions, and investigating the conditions under which they generalize. I believe this is a promising direction for future research.


\bibliography{MathWorkshop}
\bibliographystyle{plain}

\newpage
\appendix

\section*{Appendix}

\section{Additional results}

\subsection{Out-of-distribution generalization, symmetric matrice inversion}\label{app:oodinverse}

In the eigenvalue and diagonalization tasks, out-of-distribution (ood) experiments indicate that the most robust models are trained on ensembles of matrices with long-tailed eigenvalue distributions (Laplace and Gaussian). This may suggest that ood generalization happens when models are trained on datasets that contain more ``edge cases'' for this specific problem -- large absolute eigenvalues, here. This would make the choice of a good (i.e. robust) training set a problem-specific issue. 

To test this hypothesis, I experiment with the inversion of symmetric matrices.  As discussed in section~\ref{sec:failures}, the ``edge cases'' for this task are matrices with large condition numbers -- the ratio of the largest and smallest absolute eigenvalues in this particular case. If the ``edge case'' hypothesis were true, we would expect distributions with a larger range of condition numbers to generalize best. 
Table~\ref{tab:condnumbers} provides statistics about the distribution of condition numbers in our seven training and test sets. Since the uniform distribution has smaller (and less variable) condition numbers, we should expect it to generalize worst. On the other hand, the Laplace and the Marchenko-Pastur, having a broad range of condition numbers, should generalize out of distribution.

\begin{table}[h]
    \small
    \centering
    \begin{tabular}{lccc}
        \toprule
        \small
        & Median &  Third quartile & 90th percentile\\
        \midrule
        Semi-circle & 9.4 & 20.4 & 52.0 \\
        Uniform & 6.3 & 14.8 & 38.9 \\
        Gaussian &  9.0&  21.2 &  57.4  \\
        Laplace &   14.1 &  34.5 &  99.5 \\
        \midrule
        abs-semicircle & 9.5 & 20.6 & 51.7 \\
        abs-Laplace&  14.3 & 35.4 & 98.3 \\
        Marchenko-Pastur & 190 & 885 & 5293\\ 
        
       \bottomrule
    \end{tabular}
    \smallskip
    \caption{\small \textbf{Distribution of condition numbers.} On a set of 10000 randomly generated  5x5 symmetric matrices.}
    \label{tab:condnumbers}
\end{table}

Table~\ref{tab:syminverse55} presents results for $5\times 5$ symmetric matrices. As in previous experiments, models trained on positive matrices only generalize to positive test sets (the reverse being false). Models trained on the uniform set, which has the smallest condition numbers, generalize just as well as the Gaussian and Laplace models, which have the largest condition numbers. This invalidates our hypothesis. We also note that while matrix inversion is only loosely related to eigenvalues and their distribution, the Laplace model performs best on this task as well. This result needs to be confirmed, but it does suggest that certain ensembles of matrices (Laplace and Gaussian) are robust for several tasks of linear algebra. 

\begin{table}[h]
    \small
    \centering
    \begin{tabular}{lccccccc}
        \toprule
        \small
        & Semi-circle &  Uniform & Gaussian & Laplace & abs-sc & abs-Lapl& Marchenko\\
        \midrule
        Semi-circle & 81& 18 & 25 & 26 & 1 & 17 & 0 \\
        Uniform & 67 & 76 & 63 & 45 & 76 & 50 & 2 \\
        Gaussian &  62 &  72 & 63 &  45 &  71 & 51 & 5 \\
        Laplace &   65 &  75 &  65 &  49 & 76 & 58 & 7 \\
        \midrule
        Abs-semicircle & 0 & 2 & 2 & 2 & 84 & 59 & 5 \\
        Abs-Laplace&  0 & 3 & 2 & 2 & 87 & 75 & 17 \\
        Marchenko-Pastur & 0 & 3 & 3 & 2 & 85 & 66 & 16\\ 
        
       \bottomrule
    \end{tabular}
    \smallskip
    \caption{\small \textbf{Generalization with different generators.} Inversion of 5x5 symmetric matrices. Rows are training data, columns test data.}
    \label{tab:syminverse55}
\end{table}

\subsection{Out-of-distribution results: learning speeds }\label{app:eigenvalues}

Table~\ref{tab:oodspeed} indicates the number of training samples needed for a model to achieve $99\%$ accuracy on the eigenvalue task. On both the semi-circle and positive test sets, Gaussian and Laplace models are more data effective than models trained on the test distribution. On the positive test set (eigenvalues distributed as the absolute value of a semi-cricle law), the absolute Laplace is the most data-efficient of the three models trained on positive matrices. Absolute Laplace requires about $33\%$ less examples than absolute semicircle (just like Laplace vs semi-circle in the non-positive case).

\begin{table}[h!]
    \small
    \centering
    \begin{tabular}{lcc}
        \toprule
        \small
        & Semi-circle &  Absolute Semi-circle\\
        \midrule
        Semi-circle  & 3.6 & - \\
        Uniform & - & - \\
        Gaussian & 2.4 & 2.1  \\
        Laplace & 2.7 & 2.4  \\
        \midrule
        Absolute semi-circle & - & 4.5 \\
        Absolute Laplace & - & 3.9 \\
        Marchenko-Pastur & - & 7.5 \\
       \bottomrule
    \end{tabular}
    \smallskip
    \caption{\small \textbf{Learning speed of different generators.} Millions of examples to compute the eigenvalues of 5x5 matrices to 99\% accuracy. Rows are the training distributions, columns the test distributions.}
    \label{tab:oodspeed}
\end{table}

Finally, Table~\ref{tab:oodspeed2} indicates the sample size needed to achieve $85\%$ accuracy when diagonalizing $5\times 5$ matrices. Models need about ten times more data than for the eigenvalue task, but the advantage of models trained on non-compact eigenvalue distributions (Laplace and Gaussian) remains.

\begin{table}[h!]
    \small
    \centering
    \begin{tabular}{lcc}
        \toprule
        \small
        & Semi-circle \\
        \midrule
        Semi-circle  & 49.5 \\
        Uniform & 68.4  \\
        Gaussian & 45.3  \\
        Laplace & 36.9  \\
       \bottomrule
    \end{tabular}
    \smallskip
    \caption{\small \textbf{Learning speed of different generators.} Millions of examples to compute the eigenvectors of 5x5 matrices to 90\% accuracy.}
    \label{tab:oodspeed2}
\end{table}

\section{Architecture, training parameters and data sets}\label{app:architecture}

\subsection{Architecture and training}

All models used in this work are sequence-to-sequence transformers \cite{transformer17}. The models used to predict eigenvectors, in sections~\ref{sec:explain} and \ref{sec:failures}, have 6 layers in the encoder and one in the decoder, 512 dimensions and 8 attention heads. Their input are encoded with the FP15 scheme (one token per coefficient), and their output with the P1000 (three tokens, sign, mantissa in base 1000, and exponent). The ``half-trained'' model with $70\%$ accuracy used P1000 for the input and output. The model used for matrix inversion in section~\ref{sec:failures} has the same architecture as in \cite{LAWT}: 6 layers, 516 dimensions and 12 attention heads in the encoder, and 1 layer, 512 dimensions and 8 heads in the decoder. It uses FP15 for its input, and P1000 for its output. In out-of-distribution experiments, models have 6 layers in the encoder and 1 in the decoder; and either P1000 in the encoder and decoder or FP15 in the encoder and P1000 in the decoder.

Models are trained to minimize the cross-entropy between their prediction and the correct solution, encoded as sequences. They use the Adam optimiser \cite{kingma2014adam}, on batches of 64 examples, with a learning rate of 0.0001, a linear warmup phase of 10000 optimisation steps, and cosine scheduling with a period of 4000000 \cite{loshchilov2016sgdr}.

\subsection{Data sets}\label{app:generation}

The training and test data for the interpretability and failure experiments (sections~\ref{sec:explain} and~\ref{sec:failures}) are generated as in \cite{LAWT}. All matrices have independent, identically distributed (iid) coefficients, sampled from a uniform law over $[-10,10]$. In out-of-distribution experiments (section~\ref{sec:ood}), I generate symmetric matrices with iid Gaussian coefficients, with standard deviation $10/\sqrt{3}$ (same as the uniform law over $[-10,10]$). For $n \times n$ matrices, Gaussian coefficients guarantee that matrix eigenvectors are uniformly distributed in all directions of $\mathbb{R}^n$ . Since their coefficients are iid, these are Wigner matrices, and their eigenvalues are distributed according to a semi-circle law \cite{Meh2004}. To generate uniform, Gaussian and Laplace distributed matrices, I decompose $M$ into their eigenvalues $\Lambda$ and eigenvectors $H$, replace the eigenvalues by $\Lambda_2$, sampled from another distribution, and reassemble $M = H \Lambda_2 H^T$. I take the absolute values of $\Lambda$ for the abs-semicircle distribution, and those of $\Lambda_2$ for the abs-Laplace. For Marchenko-Pastur distribution, I sample a matrix $N$ with Gaussian iid coefficient, with standard deviation $\sqrt{10 / \sqrt{3}}$, and compute $M=N^T N$. All matrices are encoded using the P1000 and FP15 schemes from \cite{LAWT}. 

\section{Related works}

This paper builds on~\cite{LAWT}, which introduces the experiments, and provides initial results on out-of-distribution (OOD) generalization for the eigenvalues of $5\times 5$ matrices. I introduce a new task, inversion of symmetric matrices, conduct experiments on model failures, and expand the OOD results to larger matrices, and to two new tasks: diagonalization and matrix inversion. 

The importance of data generators in math transformers was first stressed by Lample and Charton \cite{lample2019deep}. When performing symbolic integration, they noticed that models trained on data generated by differenciating random functions performed badly on test examples generated by integrating random functions (and vice versa). Welleck et al. \cite{welleck2022symbolic} provides additional results on the lack of robustness of models trained to compute integrals. 

Yehuda et al.~\cite{yehuda20a} explore the theoretical limitations of models trained from synthetic mathematical data. They argue that model performance is limited by the training data: which instances of the problem the generator can provide. We believe our results might stand as a counter-example:  if ``long range'' out-of-distribution is possible (as suggested by our experiments), then it might be possible to solve hard instances of a problem, with a model trained on solvable instances.

\end{document}